\documentclass[10pt,twocolumn,letterpaper]{article}

\usepackage{cvpr}
\usepackage{times}
\usepackage{epsfig}
\usepackage{graphicx}
\usepackage{amsmath}
\usepackage{amssymb}
\usepackage{bm}
\usepackage{algorithm}
\usepackage{algorithmic}
\usepackage{enumerate}
\usepackage{subfigure}
\usepackage[skip=0.05\baselineskip]{caption}

\usepackage[pagebackref=true,breaklinks=true,letterpaper=true,colorlinks,bookmarks=false]{hyperref}

\cvprfinalcopy % *** Uncomment this line for the final submission

\def\cvprPaperID{64} % *** Enter the CVPR Paper ID here

\ifcvprfinal\fi
\begin{document}

%%%%%%%%% TITLE
\title{Towards Open-Set Identity Preserving Face Synthesis}

\author{Jianmin Bao$^{1}$, \quad Dong Chen$^{2}$, \quad Fang Wen$^{2}$, \quad Houqiang Li$^{1}$, \quad Gang Hua$^{2}$\\
	$^{1}$University of Science and Technology of China \qquad $^{2}$Microsoft Research\\
	{\tt\small jmbao@mail.ustc.edu.cn}  \quad{\tt\small \{doch, fangwen, ganghua\}@microsoft.com}  \quad{\tt\small lihq@ustc.edu.cn} }

\maketitle
%\thispagestyle{empty}

%%%%%%%%% ABSTRACT
\begin{abstract}

We propose a framework based on Generative Adversarial Networks to disentangle the identity and attributes of faces, such that we can conveniently recombine different identities and attributes for identity preserving face synthesis in open domains.
Previous identity preserving face synthesis processes are largely confined to synthesizing faces with known identities that are already in the training dataset.
To synthesize a face with identity outside the training dataset, our framework requires one input image of that subject to produce an identity vector, and any other input face image to extract an attribute vector capturing, {\em e.g.}, pose, emotion, illumination, and even the background. 
We then recombine the identity vector and the attribute vector to synthesize a new face of the subject with the extracted attribute.
Our proposed framework does not need to annotate the attributes of faces in any way.
It is trained with an asymmetric loss function to better preserve the identity and stabilize the training process.
It can also effectively leverage large amounts of unlabeled training face images to further improve the fidelity of the synthesized faces for subjects that are not presented in the labeled training face dataset.
Our experiments demonstrate the efficacy of the proposed framework.
We also present its usage in a much broader set of applications including face frontalization, face attribute morphing, and face adversarial example detection.
\end{abstract}

%%%%%%%%% BODY TEXT

\section{Introduction}

\begin{figure}
	\centering
	\includegraphics[width=0.8\columnwidth]{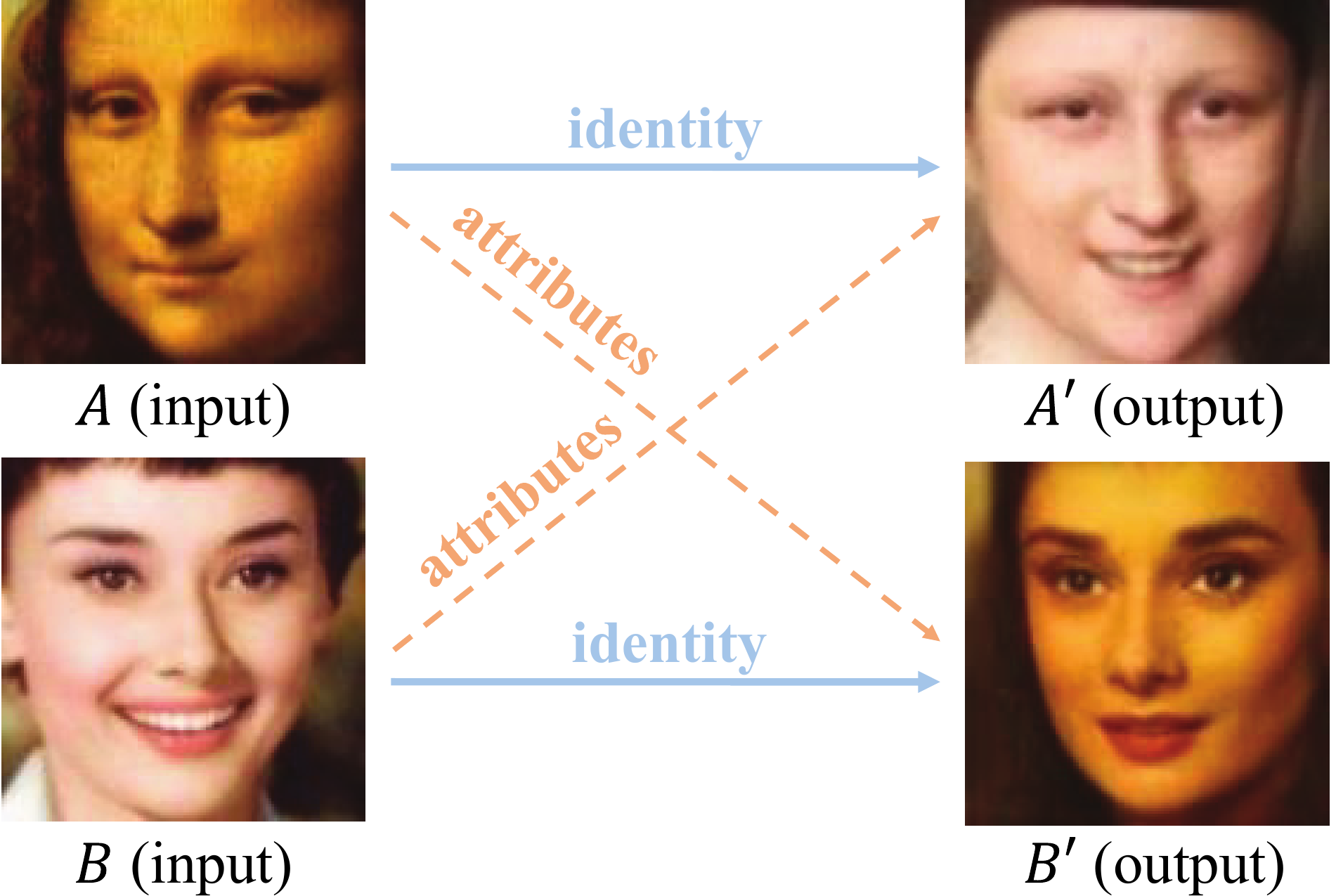}\\
	\caption{Our method can disentangle identity and attributes from a single face image. With the extracted identity and attributes from two input images $A$ and $B$, we can generate two new face images $A'$ and $B'$ by recombining the identities and attributes. }\label{fig:intro}
    \vspace{-0.5cm}
\end{figure}

Realistic face image synthesis has many real-world applications, such as face super-resolution, frontalization, and morphing, among others. With the emergence of deep generative models, such as the Generative Adversarial Networks (GAN)~\cite{goodfellow2014generative} and the Variational Auto-encoder~\cite{kingma2013auto}, we have made tremendous progress in building deep networks for synthesizing realistic faces~\cite{yan2015attribute2image, larsen2015autoencoding, tran2017disentangled, larsen2015autoencoding}. However, identity preserving face synthesis remains a challenge, especially when the identity of the face is not presented among the training face images.

%Face image synthesis is one of the most widely studied topics in computer vision.  It has many applications, such as face super resolution, face frontalization and face attribution manipulation. Recently, great progress has been achieved with Generative Adversarial Networks (GAN)~\cite{goodfellow2014generative} and following works~\cite{yan2015attribute2image, larsen2015autoencoding, tran2017disentangled, larsen2015autoencoding}. They focus on building a face generative model to capture the underlying holistic face data distribution. However, identity preserving face synthesis is still a challenge. 

Many previous works have attempted to synthesize face images of a specific person. For example, TP-GAN~\cite{huang2017beyond} and FF-GAN~\cite{yin2017towards} attempt to synthesize the frontal view of a face from a single face image. DR-GAN~\cite{tran2017disentangled} can change the pose of an input face image. However, these methods can only manipulate limited types of attributes, such as poses. These methods also require full annotation of attributes for training the models. More recent work, such as CVAE-GAN~\cite{bao2017cvae}, can produce a variety of attribute changes. Nevertheless, it is not able to synthesize a face with an identity outside the training dataset.

% the other is that visual attributes editing often cause face images losing identites info. 3D morphable face models get a success in manipulating face geometry and texture in low-dimensional linear manifolds. But the morphable face models are limited, it can't handle a much wider range faces.

In view of the limitation of CVAE-GAN, we attempt to solve the problem of open-set identity preserving face synthesis. Our goal is to be able to synthesize face images with any specific identity, no matter if the identity is presented in the training dataset or not. To synthesize a face with an identity outside the training dataset, we require one input image of that subject to prodce an identity vector, and any other input face image to extract an attribute vector capturing, {\em e.g.}, pose, emotion, illumination, and even background. We then combine the identity vector and the attribute vector to synthesize a new face of the subject with the extracted attribute.

To this end, we propose a framework based on Generative Adversarial Networks to disentangle identity and attributes given a face image, and recombine different identities and attributes for identity preserving face synthesis. As shown in Figure~\ref{fig:pipeline}, our framework has five parts: 1) an identity encoder network $I$ to encode the identities of subjects; 2) an attribute encoder network $E$ to extract attributes of any given face image; 3) a generator network $G$ to synthesize a face image from a combined input of identity and attributes; 4) a classification network $C$ to preserve the identity of the generated face; and 5) a discriminate network $D$ to distinguish real and generated examples. These five parts are trained end-to-end.

In this framework, we propose a new, simple yet elegant way to extract the attributes of an input face image. The proposed framework does not need to annotate the attributes of the faces at all. We use two loss functions: 1) a reconstruction loss of the attribute image, and 2) a $KL$ divergence loss defined on the attribute vector. These functions enforce that network $A$ extracts the attribute information. 

We take full advantage of recent advancements in face recognition, and use the softmax loss on top of network $I$ to encode the identity into an attribute independent vector representation. Therefore, in order to reconstruct the input, network $A$ is forced to extract the attribute information. Meanwhile, we add a $KL$ divergence loss to regularize the attribute vector, such that it dose not contain identity information.

Inspired by the CVAE-GAN~\cite{bao2017cvae}, we adopt a new asymmetric loss function. More specifically, we adopt a cross-entropy loss when training the discriminative network $D$, and the classification network $C$, and use a pairwise feature matching loss when updating the generative network $G$. This does a better job of preserving the identity while stabilizing the training process.

Last but not least, the proposed framework enables us to effectively leverage large amounts of unlabeled training face images to further improve the fidelity of the synthesized faces for subjects that are not presented in the labeled training face dataset. These unlabeled data can increase intra-class and inter-class variation of the face distributions, and hence improve the diversity of the synthesized faces. As a result, the generated faces present larger changes in pose and expression.

% Since we use one image to offer the identity and other to offer the attribute. so we need to disentangle identity representation and attribute representation from a face image. For disentangling the identity feature, by the achievements of face recognition, we use a face classification model to encoder the identity representation. For disentangling the attributes feature, since attributes is unlabeled, therefore, we use a fully-unsupervised method. Consider the identity input and attribute input is the same face, so the model do a reconstruction task, with an appropriate prior set the attribute distribution, the reconstruction task will force the attribute encoder to disentangle the attribute.

% On the other hand, we want to generate idenity-preserving face images. So we add a face classification network after the generative network. To keep the identities of the generated images. we use asymmetric training method which means we use the softmax loss for classification network but perceptual loss for the generative network. More specially, we want to generate realistic face images, we also add a discriminator network after the generative network and also use the asymmetric training strategy.

% Open-set face generation is still very challenging during face generation tasks. To solve this, we use a unsupervised method. Beside the labeled identity data, we further add many unlabeled face data to train our framework. This help our generative model to improve the image quality of people who are not in the training dataset. And handle the large variations and emotions.

Our experiments demonstrate the efficacy of the proposed framework. We also show that our model can be applied to other tasks, such as face frontalization, face attribute morphing and face adversarial examples detection.

\section{Related Work}
We briefly summarize the most related works, ranging from general literature of deep generative models to more specific models on face synthesis.

There have been many recent development on deep generative modeling, including deterministic generative models~\cite{dosovitskiy2016learning, reed2015deep}, Generative Adversarial Networks (GAN)~\cite{goodfellow2014generative,radford2015unsupervised}, Variational Auto-encoders (VAE)~\cite{kingma2013auto,rezende2014stochastic}, and autoregression networks~\cite{larochelle2011neural}, to list a few. Among them, the most popular one is perhaps GANs~\cite{goodfellow2014generative,radford2015unsupervised}.

Many variants of GANs have been proposed to improve the stability of training process~\cite{salimans2016improved, arjovsky2017wasserstein, berthelot2017began}. Meanwhile, there are also many works that have added condition information to the generative network and the discriminative network for conditional image synthesis. The condition information could be a discrete label~\cite{mirza2014conditional, chen2016infogan, bao2017cvae, larsen2015autoencoding, odena2016conditional}, a reference image~\cite{liu2017unsupervised, dong2017semantic, yoo2016pixel, chen2017stylebank, chen2017coherent}, or even a text sentence~\cite{reed2016generative, zhang2016stackgan, dong2017semantic}.
%some works focused on image generation conditioned on a given discrete label~\cite{mirza2014conditional, chen2016infogan, bao2017cvae, larsen2015autoencoding}, while some other works synthesize images conditioned on input images~\cite{liu2017unsupervised, dong2017semantic, yoo2016pixel}. Many other works also focused on synthesizing realistic images from a given text description~\cite{reed2016generative, zhang2016stackgan, dong2017semantic}. 

Due to its abundant useful applications, the face is a major focus of image generation. Antipov {\em et al.}~\cite{antipov2017face} propose a variant of GANs for face aging. 
%Also, there is some work focus on identity preserving face generation. 
Li {\em et al.}~\cite{li2016convolutional} propose a method for attribute-driven and identity-preserving face generation. However, the attribute is only limited to some simple ones. TP-GAN~\cite{huang2017beyond} adopt a two-pathway generative network to synthesize frontal faces. Both DR-GAN and TP-GAN obtained impressive results on face frontalization, but they need to explicitly label the frontal faces.

Prior work also explores disentangled representation learning. For example, the DC-IGN~\cite{kulkarni2015deep} uses a variational auto-encoder based method to learn the disentangled presentation. However, DC-IGN needs to fix one attribute in one batch training, which also needs explicit annotations of the attributes. Luan {\em et al.}~\cite{tran2017disentangled} proposed DR-GAN to learn a generative and discriminative representation, which explicitly disentangles the pose leveraging the pose annotations.

In contrast, this paper proposes an Identity Preserving Generative Adversarial Network framework, which does not require any attribute annotations. This framework disentangles the identity and attributes representations, and then uses different recombinations of representations for identity preserving face synthesis. This disentaglement allows us to synthesize faces with identities outside what is presented in the training datasets. This addresses a serious limitation of a previous deep generative model-based identity preserving face synthesis method~\cite{bao2017cvae}. It simply can not generate faces of identities outside the training dataset.

\section{Identity Preserving GANs}
\begin{figure*}[t]
	\centering
	\includegraphics[width=2.0\columnwidth]{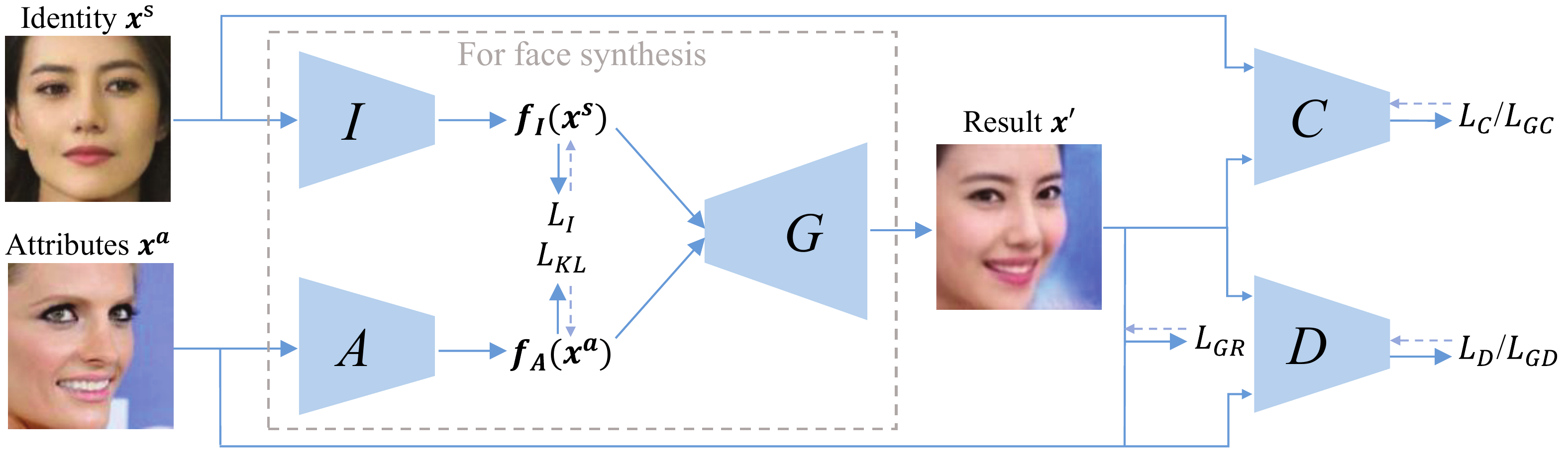}\\
	\caption{Framework overview: we train a face synthesis network to disentangle identity and attributes from face images and recombine them differently to produce result $\bm{x'}$, which uses the identity of $\bm{x^s}$ and attributes of $\bm{x^a}$. The input/output are drawn with solid lines. The loss functions are drawn with the dashed lines.}\label{fig:pipeline}
    \vspace{-0.5cm}
\end{figure*}

In this section, we introduce our face synthesis networks. To synthesize a face of any specific identity, our framework requires two input images, {\em i.e.}, one input image $\bm{x^s}$ of a certain subject identity, and another input image $\bm{x^a}$ to extract the attributes, {\em e.g.}, pose, emotion, illumination, and even background. Our network synthesizes a new face image $\bm{x'}$ of the subject with the extracted attributes.

As show in Figure~\ref{fig:pipeline}, our framework is based on Generative Adversarial Networks. It contains five parts: 1) the identity encoder network $I$; 2) the attributes encoder network $A$; 3) the generative network $G$; 4) the classification network $C$; and 5) the discriminative network $D$. The function of the network $I$ is to extract the identity vector $\bm{f_I}(\bm{x^s})$ from the subject image $\bm{x^s}$. The network $A$ is adopted to extract the attribute vector $\bm{f_A}(\bm{x^a})$ of the attributes image $\bm{x^a}$. 

The network $G$ generates a new face image $\bm{x'}$ using the combined identity vector and attribute vector $[\bm{f_I}(\bm{x^s})^T, \bm{f_A}(\bm{x^a})^T]^T$. The network $C$ and the network $D$ are only included in the training phase. The network $C$ is used to preserve the identity by measuring the posterior probability $P(c|\bm{x^s})$, where $c$ is the subject identity of $\bm{x^s}$. The discriminative network $D$ distinguishes between real and generated examples.

As we do not want to annotate the attributes, extracting the attribute vector from a face image poses a challenge. In the following sections, we first introduce our method of disentangling the identity vector and the attribute vector from a face image in Section~\ref{sec:Disentangle}. Then, in Section~\ref{sec:Loss function}, we introduce an asymmetric training method to generate identity-preserving and realistic face images, and to make the training process more stable. 

In Section~\ref{sec:Semi-Supervised Learning}, in order to further improve the fidelity of the synthesized faces for subjects that are not presented in the labeled training face dataset, we use an unsupervised learning method with a large amount of unlabeled training images. Finally, in Section~\ref{sec:Overall Objective Function}, we analyze the objective function of the proposed method and provide the implementation details of the training pipeline.

\subsection{Disentanglement of Identity and Attributes}
\label{sec:Disentangle}
In this section, we introduce the technical details of disentangling the identity vector and the attribute vector using network $I$ and network $A$, respectively. In our training data, we only have the annotation of the identity of each face, without any annotation of the attribute information. This is because face images with category annotations are relatively easy to obtain. Many large datasets are publically available, such as the FaceScrub~\cite{ng2014data}, CASIA-WebFace~\cite{yi2014learning} and MS-Celeb-1M~\cite{guo2016ms} datasets. However, the annotation of attributes is often more difficult, and sometimes even impossible, such for illumination and the background.

Extracting the identity vector is relatively straightforward. Here, we take full advantage of recent improvement in face recognition. Given a set of face images with the identity annotation $\{\bm{x}_i^s, c_i\}$, we use the softmax loss for training network $I$ to perform face classification task. Therefore, the same individuals have approximately the same feature which can be used as the identity vector. Formally, the loss function of the network $I$ is
\begin{equation}
\label{eqn:L_I}
\mathcal{L_I} =  -\mathbb{E}_{\bm{x} \thicksim P_{r}}[\log{P(c|\bm{x^s})}],
\end{equation}
where $P(c|\bm{x^s})$ represents the probability of $\bm{x^s}$ having identity $c$. Then, we use the response of the last pooling layer of $I$ as the identity vector.

In order to train network $A$ in a fully unsupervised manner to extract the attribute vector. We propose a new, simple yet elegant way to extract the attributes of each face. In the training process, we consider two loss functions: reconstruction loss and $KL$ divergence loss.

% situations: one is the source image $\bm{x^s}$ is equal to target image $\bm{x^a}$, we called this the reconstruction process, the other is that the source image $\bm{x^s}$ is not equal to target image $\bm{x^a}$ which we called a Transformation process. we will introduce our method on how to learn the attributes representation $f_A(\bm{x^a})$ under these two situations.

\noindent \textbf{Reconstruction loss.} Here we have two situations: whether subject image $\bm{x^s}$ is the same as attribute image $\bm{x^a}$ or not. In both cases, we require the result image $\bm{x}'$ to reconstruct attribute image $\bm{x^a}$, but with different loss weight. Formally, the reconstruction loss is
\begin{equation}
\vspace{-0.1cm}
\label{eqn:L_GR}
\mathcal{L}_{GR} = \left\{
\begin{aligned}
\frac{1}{2}||\bm{x^a}- \bm{x}'||_2^2&\qquad\text{if $\bm{x^s} = \bm{x^a}$}\\
\frac{\lambda}{2}||\bm{x^a}- \bm{x}'||_2^2&\qquad\text{otherwise}
\end{aligned}
\right.,
\vspace{-0.2cm}
\end{equation}
where $\lambda$ is the reconstruction loss weight. Next, we analyze the reconstruction loss under these two situations.

When subject image $\bm{x^s}$ is the same as attribute image $\bm{x^a}$, output $\bm{x}'$ must to be the same as $\bm{x^s}$ or $\bm{x^a}$. Supposing there are various face images of an identity, the identity vector $\bm{f_I}(\bm{x})$ is almost the same for all samples. But the reconstruction using the $\bm{f_I}(\bm{x})$ and $\bm{f_A}(\bm{x})$ of different samples are all different. Therefore, the reconstruction loss will force the attribute encoder network $A$ to learn different attributes representation $\bm{f_A}(\bm{x})$. 

When subject image $\bm{x^s}$ and attribute image $\bm{x^a}$ are different, we do not know exactly what the reconstructed result should look like; but we can expect the reconstruction to be approximately similar to the attribute image $\bm{x^a}$, such as the background, overall illumination, and pose. Therefore, we adopt a raw pixel reconstruction loss with a relatively small weight to maintain the attributes. We set $\lambda = 0.1$ in our experiments. As expected, a large $\lambda$ causes poor results. we further demonstrate this in the supplementary material due to space limits.

\noindent \textbf{$KL$ divergence loss.} To help the attributes encoder network learn better representations, we also add a $KL$ divergence loss to regularize the attribute vector with an appropriate prior $P(\bm{z}) \thicksim N(0, 1)$. The $KL$ divergence loss will limit the distribution range of the attribute vector, such that it dose not contain much identity information, if at all. For each input face image, the network $A$ outputs the mean $\bm{\mu}$ and covariance of the latent vector. We use the $KL$ divergence loss to reduce the gap between the prior $P(\bm{z})$ and the learned distributions, {\em i.e.},
\begin{equation}
\vspace{-0.1cm}
\mathcal{L}_{KL} = \frac{1}{2}(\bm{\mu}^T\bm{\mu}+\sum_{j-1}^{J}(\exp (\bm{\epsilon})-
\bm{\epsilon}-1)),
\vspace{-0.2cm}
\end{equation}
where $j$ denotes the $j$-th element of vector $\bm{\epsilon}$. Similar to the Variational Auto-encoder~\cite{kingma2013auto}, we sample the attribute vector using $\bm{z} = \bm{\mu}+\bm{r} \odot exp(\bm{\epsilon})$ in the training phase, where $\bm{r} \thicksim N(\bm{0}, \bm{I})$ is a random vector and $\odot$ represents the element-wise multiplication. 

\subsection{Asymmetric Training for Networks $G$, $C$, and $D$}
\label{sec:Loss function}

After extracting the identity vector $\bm{f_I}(\bm{x}^s)$ and attribute vector $\bm{f_A}(\bm{x}^a)$, we concatenate them in the latent space and feed the combined vector, $\bm{z} = [\bm{f_I}(\bm{x^s})^T, \bm{f_A}(\bm{x^a})^T]^T$ into the network G to synthesize a new face image. In this section, we introduce our asymmetric training method. It can generate identity-preserving and realistic face images, which also makes the training process more stable.

Similar to GANs, the generative network $G$ competes in a two-player minimax game with the discriminative network $D$. Network $D$ tries to distinguish real training data from synthesized data, while network $G$ tries to fool the network $D$. Concretely, network $D$ tries to minimize the loss function

\vspace{-0.1cm}
\begin{equation}
\label{eqn:L_D}
\mathcal{L}_{D} =  -\mathbb{E}_{\bm{x} \thicksim P_{r}}[\mathrm{log} D(\bm{x^a})] - \mathbb{E}_{\bm{z} \thicksim P_{\bm{z}}}[\mathrm{log} (1 - D(G(\bm{z}))],
\end{equation}

However, if the network $G$ directly attempts to maximize $\mathcal{L}_D$ as the traditional GAN, the training process of network $G$ will be unstable. This is because, in practice, the distributions of ``real'' and ``fake'' images may not overlap with each other, especially at the early stages of the training process. Hence, network $D$ can separate them perfectly. That is, we always have $D(\bm{x}^a)\to1$ and $D(\bm{x'})\to0$, as a result, $\mathcal{L}_D\to0$. Therefore, when updating the network $G$, the gradient $\partial \mathcal{L}_D/\partial G \to 0$. This causes gradient vanishing. Recent works~\cite{arjovsky2017towards,arjovsky2017wasserstein} also theoretically analyze the gradient vanishing problem in training GANs.

To address this problem, inspired by CVAE-GAN~\cite{bao2017cvae}, we propose a pairwise feature matching objective for the generator. To generate realistic face image quality,  we match the feature of the network $D$ of real and fake images. Let $\bm{f_D}(\bm{x})$ denote features on an intermediate layer of the discriminator, then the pairwise feature matching loss is the Euclidean distance between the feature representations, {\em i.e.},

\vspace{-0.2cm}
\begin{equation}
\label{eqn:L_GD_stat}
\mathcal{L}_{GD} =  \frac{1}{2}||\bm{f_D}(\bm{x}') - \bm{f_D}(\bm{x^a})||_2^2.
\vspace{-0.2cm}
\end{equation}
In our experiment, for simplicity, we choose the input of the last Fully Connected (FC) layer of network $D$ as the feature $\bm{f_D}$.

Meanwhile, classification network $C$ tries to classify faces of different identities, meaning it tries to minimize the loss function

\vspace{-0.2cm}
\begin{equation}
\label{eqn:L_C}
\mathcal{L_C} =  -\mathbb{E}_{\bm{x} \thicksim P_{r}}[\log{P(c|\bm{x^s})}].
\vspace{-0.2cm}
\end{equation}

\noindent In order to generate identity-preserving face images, we also use pairwise feature matching to encourage $\bm{x'}$ and $\bm{x^s}$ to have similar feature representations in network $C$. Let $\bm{f_C}(\bm{x})$ denote features produced from an intermediate layer of the classification network $C$. The feature reconstruction loss is the Euclidean distance between feature representations, {\em i.e.},

\vspace{-0.1cm}
\begin{equation}
\label{eqn:L_GC_stat}
\mathcal{L}_{GC} =  \frac{1}{2}||\bm{f_C}(\bm{x}') - \bm{f_C}(\bm{x^s})||_2^2.
\vspace{-0.2cm}
\end{equation}
Here, we choose the input of the last FC layer of network $C$ as the feature for simplicity. We also try to combine features of multiple layers, it only marginally improves the ability to preserve the identity of network $G$. network $C$ and network $I$ can share the parameters and be initialized by a pretrained face classification network to speed up the convergence.

\subsection{Unsupervised Training}
\label{sec:Semi-Supervised Learning}
Generating face images of identities which are not presented in the labeled training dataset remains a challenge. It requires the generative network to cover all intra-person and inter-person variations. Existing publically available training datasets with labeled identities are often limited by size, usually do not contain extreme poses or illuminations. In other words, they are not diverse enough.

To solve this problem, we randomly collect about $1$ million face images from flicker and Google, and use a face detector to locate the face region. These images have much larger variation, and hence are much more diverse than any existing face recognition datasets. We add these data into the training dataset, and perform an unsupervised training process to help train our generative model to better synthesize face images that do not appear in the training set.

These unlabeled images can be used either as the subject image $\bm{x^s}$ or the attribute image $\bm{x^a}$. When used as the attribute image $\bm{x^a}$, the whole training process remains unchanged. When used as the subject image $\bm{x^s}$, since it does not have a class label, we ignore the loss function $\mathcal{L}_I$ and $\mathcal{L}_C$. In other words, since networks $I$ and $C$ are fixed, we update the other parts of the end-to-end framework.

These unlabeled data can increase intra-class and inter-class variation of the face distributions, hence improving the diversity of the synthesized faces. As a result, the generated faces present larger changes in poses and expressions. We demonstrate this in the experiments.

\subsection{Overall Objective Function}
\label{sec:Overall Objective Function}

\begin{table}
	\small
	\centering
	\begin{tabular}{|c|c|c|c|c|c|} \hline
		Net& I & A & G & D & C \\
		\hline
		Loss&$\mathcal{L}_I $&$\mathcal{L}_{KL}$, $\mathcal{L}_{GR}$&$\mathcal{L}_{GR}$, $\mathcal{L}_{GC}$, $\mathcal{L}_{GD}$&$\mathcal{L}_D$&$\mathcal{L}_C$\\
		\hline
	\end{tabular}
	\caption{Networks and their related loss function.}  
    \label{tab:network_and_loss}
    \vspace{-0.6cm}
\end{table}

The final synthesis loss function is the sum of all the losses defined above in Equation~\ref{eqn:L_I} - \ref{eqn:L_GC_stat}. Although there are many loss functions, as shown in Table~\ref{tab:network_and_loss}, each network relates to only one part of the loss function. Therefore, our framework is easy to train and does not need to balance the various loss functions.

In the training phase, we separate each iteration into two steps: one step for the reconstruction process when $\bm{x^s}=\bm{x^a}$ and one step for the transformation process when $\bm{x^s} \neq \bm{x^a}$. The details of the training algorithm are described in Algorithm~\ref{alg:pipeline}. 
 
 \begin{algorithm}[t]
 	\caption{Two training process strategy. $\theta_{I}$, $\theta_{A}$, $\theta_{G}$, $\theta_{D}$, and $\theta_{C}$ are the initial parameters of networks $I$, $A$, $G$, $D$, and $C$. $iter \leftarrow 1$.}
 	\label{alg:pipeline} 	
 	\begin{algorithmic}
	    \WHILE{$\theta_{G}$ not converaged}
       		\STATE Sample ${\bm{x^s}, c}$ a batch from the dataset;
	        \IF{$iter \% 2 = 1$}
	        	\STATE // Training at reconstruction process
		        \STATE $\lambda$ $\leftarrow$ $1$
           		\STATE $\bm{x^a}$ $\leftarrow$ $\bm{x^s}$ 
			\ELSE
		    	\STATE // Training at transformation process
          		\STATE Sample ${\bm{x^a}, c}$ a batch from the dataset;
		        \STATE $\lambda$ $\leftarrow$ $0.1$
		    \ENDIF
       		\STATE $\mathcal{L}_{I}$ $\leftarrow$ $-$log($P(c|\bm{x^s})$)
            \STATE $\mathcal{L}_{C}$ $\leftarrow$ $-$log($P(c|\bm{x^s})$)
       		\STATE $\bm{f_I}(\bm{x^s})$ $\leftarrow$ $I(\bm{x^s})$; $\bm{f_A}(\bm{x^a})$ $\leftarrow$ $A(\bm{x^a})$
       		\STATE $\mathcal{L}_{KL}$ $\leftarrow$ $KL(\bm{f_A}(\bm{x^a})||P(\bm{z}))$
       		\STATE $\bm{x'}$ $\leftarrow$ $G([\bm{f_I}(\bm{x^s})^T, \bm{f_A}(\bm{x^a})^T]^T)$
            \STATE $\mathcal{L}_{D}$ $\leftarrow$ $ -$(log($D(\bm{x^a})$) + log(1- $D(\bm{x'})$))
			\STATE $\mathcal{L}_{GR}$ $\leftarrow$ $\frac{1}{2}||\bm{x^a}- \bm{x'}||_2^2$
			\STATE $\mathcal{L}_{GD}$ $\leftarrow$ $\frac{1}{2}||\bm{f_D}(\bm{x^a}) - \bm{f_D}(\bm{x'})||_2^2$
			\STATE $\mathcal{L}_{GC}$ $\leftarrow$ $\frac{1}{2}||\bm{f_C}(\bm{x^s}) - \bm{f_C}(\bm{x'})||_2^2$
            \STATE $\theta_{I}$ $\stackrel{+}{\longleftarrow}$ $-\nabla_{\theta_{I}}(\mathcal{L}_{I})$ 
            \STATE $\theta_{C}$ $\stackrel{+}{\longleftarrow}$ $-\nabla_{\theta_{C}}(\mathcal{L}_{C})$
            \STATE $\theta_{D}$ $\stackrel{+}{\longleftarrow}$ $-\nabla_{\theta_{D}}(\mathcal{L}_{D})$
            \STATE $\theta_{G}$ $\stackrel{+}{\longleftarrow}$ $-\nabla_{\theta_{G}}(\lambda\mathcal{L}_{GR} + \mathcal{L}_{GD} + \mathcal{L}_{GC})$
            \STATE $\theta_{A}$ $\stackrel{+}{\longleftarrow}$ $-\nabla_{\theta_{A}}(\lambda\mathcal{L}_{KL} +\lambda\mathcal{L}_{G})$     
            \STATE {$iter$ $\leftarrow$ $iter + 1$}
		\ENDWHILE		
		
	\end{algorithmic}
\end{algorithm}

\section{Experiments}
We use a subset of the MS-Celeb-1M~\cite{guo2016ms} dataset, which contains about $5M$ images of $80K$ celebrities for training. For each face image, we first detect the facial region with the JDA face detector~\cite{chen2014joint}, and then align and resize it to $128 \times 128$ pixels. 

For networks $I$, $C$, and $A$, we use the same VGG network~\cite{simonyan2014very} structure. Meanwhile, $I$ and $C$ share parameters in the training stage to speed up convergence. For network $G$, it is an inverse VGG structure. Its pooling layers are replaced by upsampling layers, and the convolution layers are replaced with deconvolution layers. For network $D$, we use the same discriminative network structure as the DCGAN~\cite{radford2015unsupervised}. The batch normalization~\cite{ioffe2015batch} layer is also applied after each convolution and deconvolution layer. The model is implemented using the deep learning toolbox Torch.

\subsection{Analysis of the Proposed Framework}
In this section, we perform an ablation study of our framework, sweeping loss combinations and different training strategies to understand how each component works in our framework. Both quantitative and qualitative results are reported.

We compare five variations of our framework: 1) removing the loss $\mathcal{L}_{GD}$; 2) removing the loss $\mathcal{L}_{GC}$; 3) training without the transformation training process (denoted as w/o $T$); 4) training without the unsupervised learning (denoted as w/o $U$); 5) our best model with all components. The network structure and training strategy remain the same for all settings.

% To demonstrate how the losses $\mathcal{L}_{GD}$,and $\mathcal{L}_{GC}$ work, we keep the same network structure and training strategy and remove one of the two losses to train two different models. Specially, we train a network without using the transform training process(simplified as \emph{T}). To validate the performance of unsupervised learning(simplified as \emph{U}) in our work, we also train a network use both the labeled data and the unlabed face data. 

\begin{figure}[t]
	\centering
	\includegraphics[width=0.75\columnwidth]{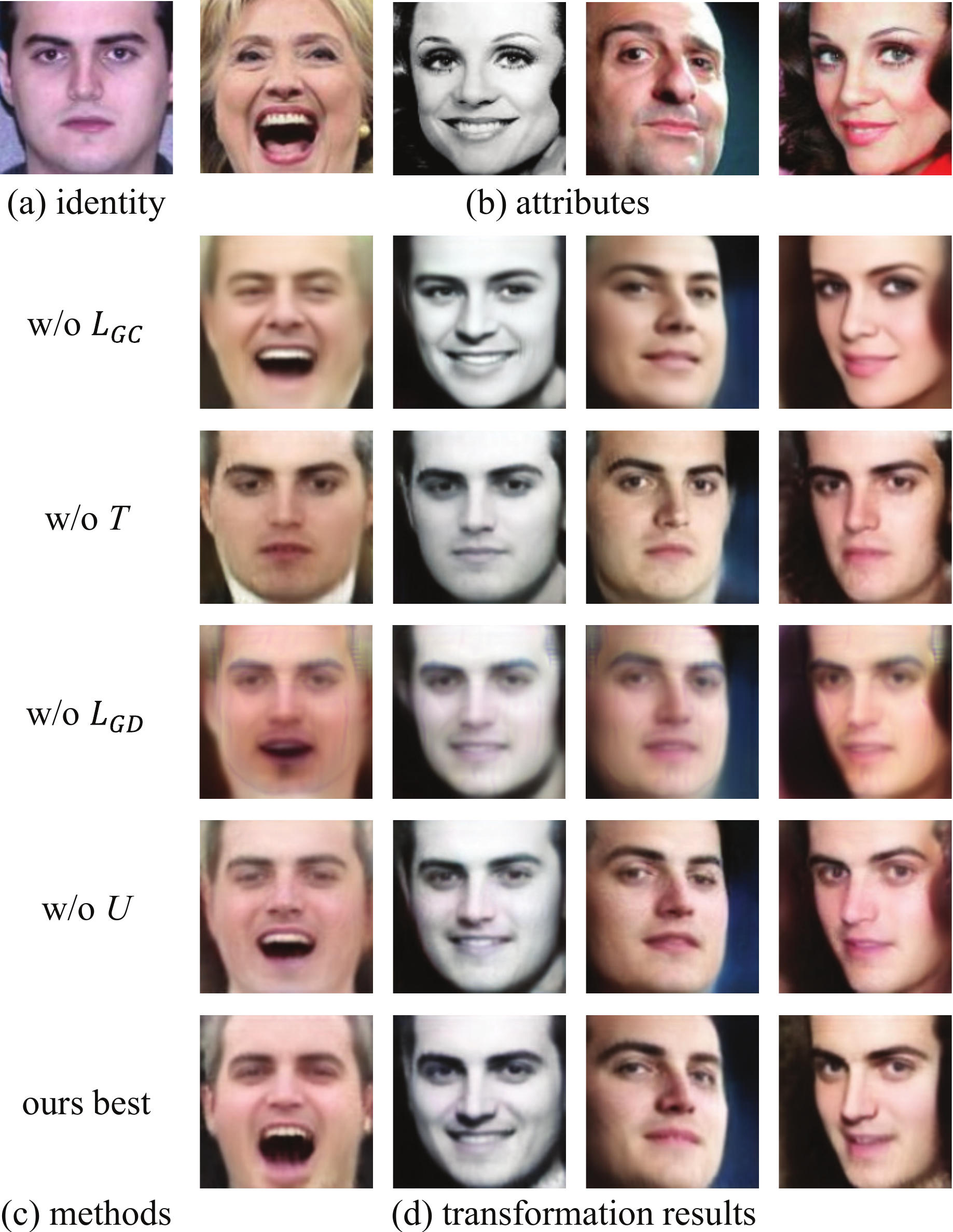}\\
	\caption{Qualitative comparison between different generative models using different loss combinations and training strategies.}\label{fig:ablation_study_qualitative}
    \vspace{-0.3cm}
\end{figure}

To quantitatively evaluate our framework, we conduct two face identification experiments to compare the performance of each setting. For identities that are appeared in datasets, We randomly pick $10K$ identities form the MS-Celeb-1M~\cite{guo2016ms} dataset, each with six photos, one for gallery and five for queries. None of these photos are in our training data. For each person, we generate $5$ images using the queries and 5 randomly selected attribute images. Then we use the generated image to find the most similar faces in the gallery, and measure the top-1 accuracy. 

For identities that are not appeared in the datasets, We use the Multi-PIE dataset. We choose 6 kinds of attributes from Multi-PIE dataset. Then for each person in the dataset, faces with one kind of attributes are set as the galleries, queries are the original face images and the generated images with the rest kinds of attributes.

\begin{table}
	\small
	\centering
	\begin{tabular}{p{1.3cm}|p{0.85cm}|p{0.65cm}|p{0.65cm}|p{0.65cm}|p{0.65cm}|p{0.65cm}}
		\hline
		Method & original data & w/o $L_{GC}$ & w/o \emph{T} & w/o $L_{GD}$ & w/o \emph{U} & ours best \\
		\hline
		Top-1 acc & 87.39 & 5.71 & 79.19 & 80.52 & 80.24 & \bf{81.11} \\  
		\hline
	\end{tabular}
%	\caption{Model comparison:  Top-1 identification rates ($\%$) on the MS-Celeb-1M~\cite{guo2016ms} dataset.}
    \caption{Model comparison:  Top-1 identification rates ($\%$) on the MS-Celeb-1M dataset.}
	\label{tab:quantitative_comparison}
	\vspace{-5mm}
\end{table}

In Table~\ref{tab:quantitative_comparison} and Table~\ref{tab:multipie_comparison}, we report the face identification top-1 accuracy of each setting. All components can improve the identity preserving capability of the framework. Among them, the $\mathcal{L}_{GC}$ contributes the most. Meanwhile, we also measure the top-1 accuracy using the real query image, Our generated images achieves comparable results.

% We choose $10K$ identities to do this experiments. First, we randomly choose one image for each identity for the gallery images. Then we randomly choose another image for each identity as the query images. Then by comparing the cosine distance of each query image to gallery images, we can get a mean top-1 accuracy. We evaluate our generative model under two settings: 1) we do a reconstruction task on each query images; 2) we choose some fixed images as the attributes, then we do the face transformation task for each query images. With these reconstructed images and transformed images, we do the face identitication task. Our evaluation model can get 86.74\% top-1 accuracy on origin query images. 

Figure~\ref{fig:ablation_study_qualitative} illustrates the qualitative results of four variants. We can observe that  removing the transformation training process cause the generated results lose attribute details, especially the emotion. Removing the loss $\mathcal{L}_{GD}$ will cause the generated image to be blurry. Removing the loss $\mathcal{L}_{GC}$, the generated samples cannot keep the identity information. The results generated by our full model achieves better results. After using unlabeled data for the unsupervised learning, our model can generate images with larger variations. For example, as the bottom left image shown in Figure~\ref{fig:ablation_study_qualitative}, the mouth can be opened wider.

\subsection{Function of $KL$ Divergence Loss}
In this section, we will verify whether the $KL$ divergence loss is able to help to remove the identity information in the attribute vector. We first train two models with and without the $KL$ divergence loss, respectively. Then, we conduct the experiment using the FaceScrub~\cite{ng2014data} dataset. We randomly split these faces into two parts, one for training and the other for validation. For each setting, we use the network $A$ to extract the attribute vector for all the faces in the Facescrub dataset. We then use a multilayer perceptron (MLP) to train a classification model to distinguish the face features of different identities in the training set.  We also test the top-1 accuracy on the validation set. 

\begin{figure}[t]
	\centering
	\includegraphics[width=1.0\columnwidth]{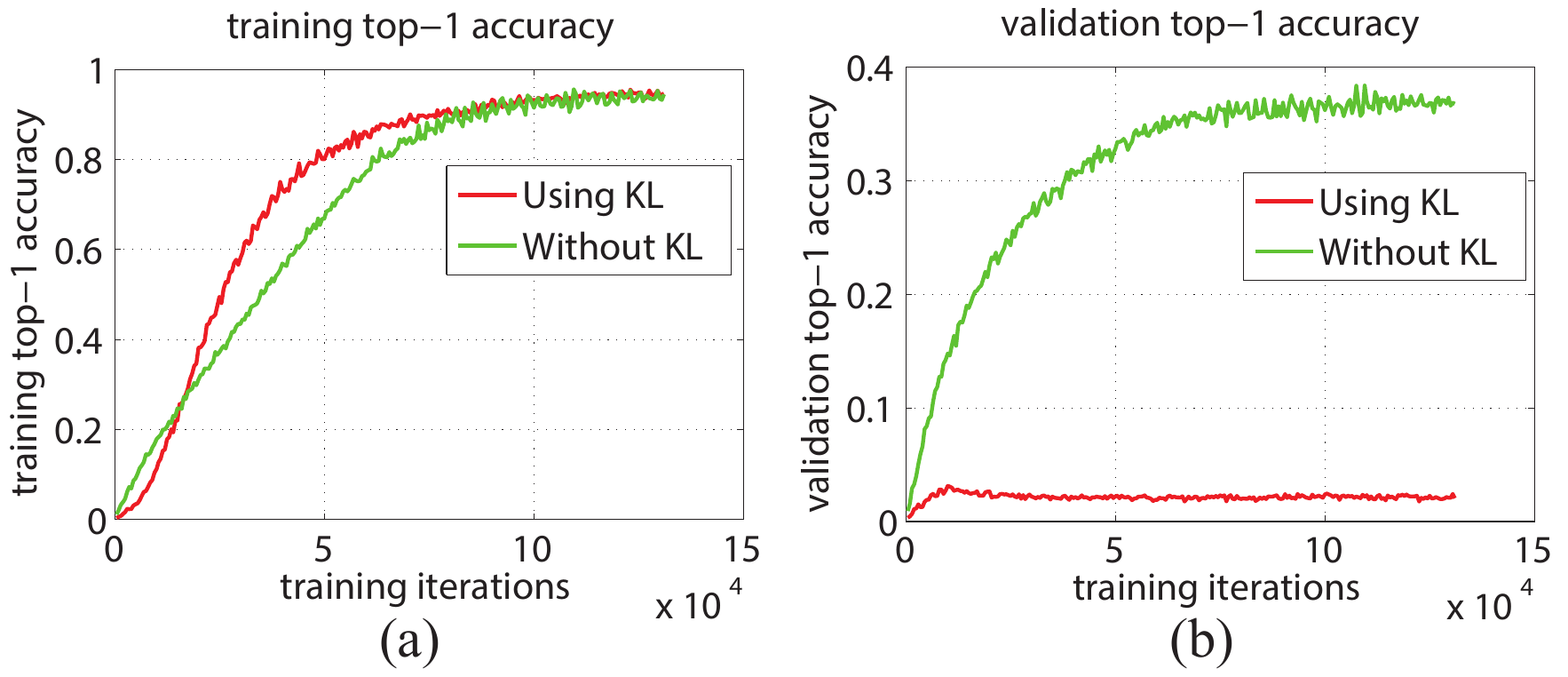}\\
	\caption{Analysis of $KL$ divergence loss. The attribute vector learned with $KL$ gets a lower top-1 accuracy.}\label{fig:KL_distribution}
    \vspace{-0.3cm}
\end{figure}

The results are presented in Figure~\ref{fig:KL_distribution}. We can see that in the validation set, the attribute vector learned using $KL$ loss has the lower top-1 accuracy, which means that it contains less identity information. It validates that the $KL$ divergence loss is able to help network $A$ to remove identity information in the attribute vector.

\begin{table}
	\small
	\centering
	\begin{tabular}{p{1.3cm}|p{0.85cm}|p{0.65cm}|p{0.65cm}|p{0.65cm}|p{0.65cm}|p{0.65cm}}
		\hline
		Method & original data & w/o $L_{GC}$ & w/o \emph{T} & w/o $L_{GD}$ & w/o \emph{U} & ours best \\
		\hline
		Top-1 acc & 97.47 & 11.76 & 95.47 & 95.53 & 96.41 & \bf{96.80} \\  
		\hline
	\end{tabular}
	\caption{Model comparison: Top-1 identification rates ($\%$) on the Multi-PIE dataset.}
	\label{tab:multipie_comparison}
	\vspace{-5mm}
\end{table}

\subsection{Face Attributes Transformation}
\begin{figure*}[t]
	\centering
	\includegraphics[width=2.0\columnwidth]{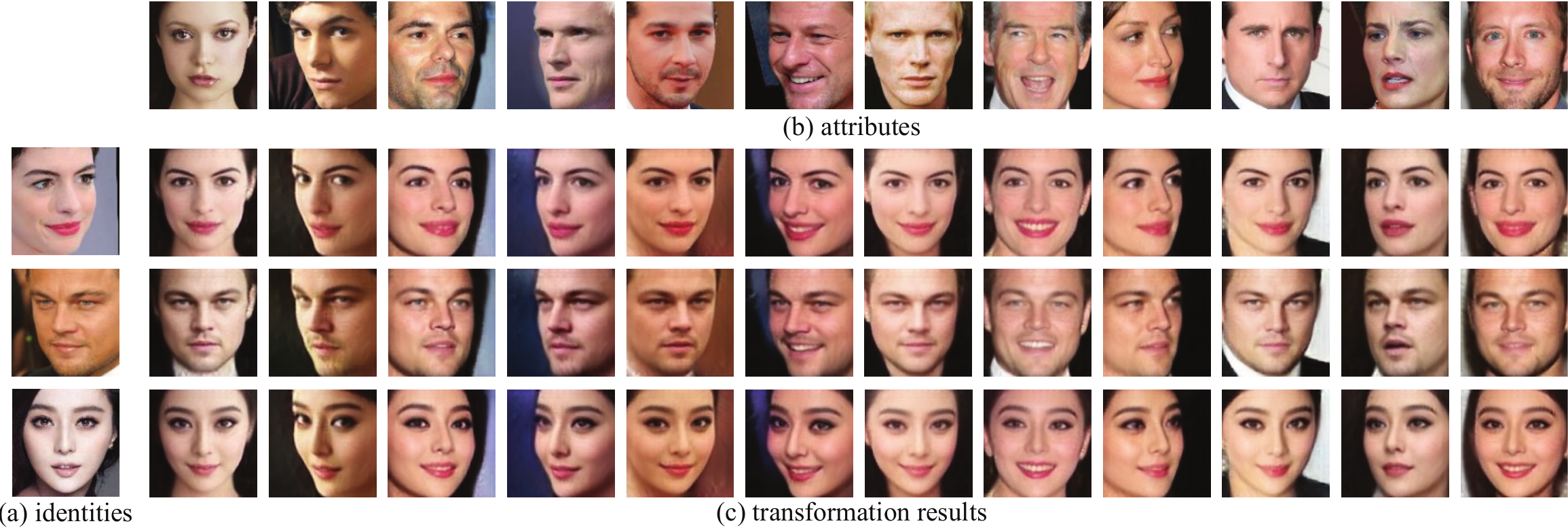}\\
	\caption{Face synthesis results using the identities that appear in training dataset and randomly chosen images as attributes.}\label{fig:seen_face_attributes_transfromation}
    \vspace{-0.3cm}
\end{figure*}

This section presents the results of face attribute transformation. The goal of face attribute transformation is to generate an image $\bm{x'}$ that combines the identity of a input face $\bm{x^s}$ and the attributes of another input face $\bm{x^a}$. To demonstrate that our framework has the ability to generate faces with identities that do not exist in the training dataset, we conduct two experiments: generating face images of identities inside or outside the training dataset, respectively.

\begin{figure*}[t]
	\centering
	\includegraphics[width=2.0\columnwidth]{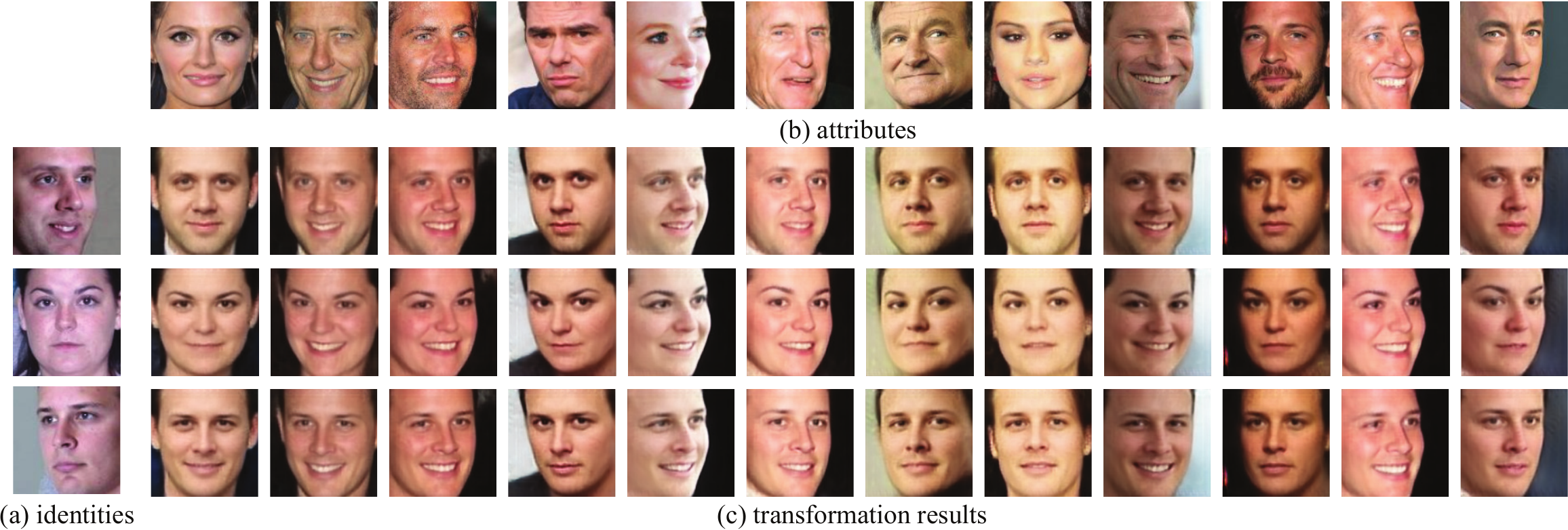}\\
	\caption{Face synthesis results using the zero-shot identities and randomly chosen images as attributes.}\label{fig:unseen_face_attributes_transfromation}
    \vspace{-0.5cm}
\end{figure*}

\begin{figure}[t]
	\centering
	\subfigure[LFW]{
		\begin{minipage}[b]{0.13\linewidth}
			\centering
			\includegraphics[width=1\linewidth]{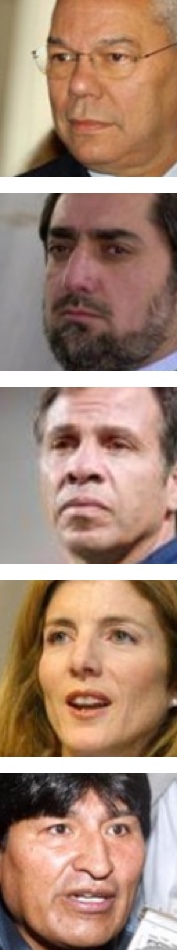}\vspace{0.03\linewidth}
	\end{minipage}}
	\subfigure[Ours]{
		\begin{minipage}[b]{0.13\linewidth}
			\centering
			\includegraphics[width=1\linewidth]{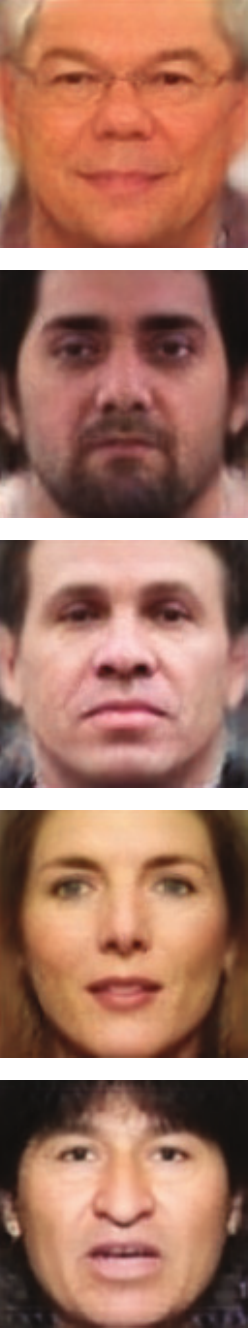}\vspace{0.03\linewidth}
	\end{minipage}}
	\subfigure[~\cite{huang2017beyond}]{
		\begin{minipage}[b]{0.13\linewidth}
			\centering
			\includegraphics[width=1\linewidth]{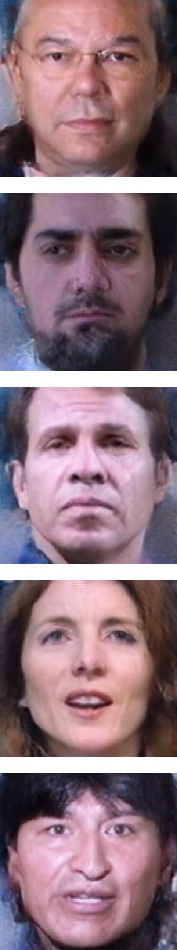}\vspace{0.03\linewidth}
	\end{minipage}}
	\subfigure[~\cite{zhu2015high}]{
		\begin{minipage}[b]{0.13\linewidth}
			\centering
			\includegraphics[width=1\linewidth]{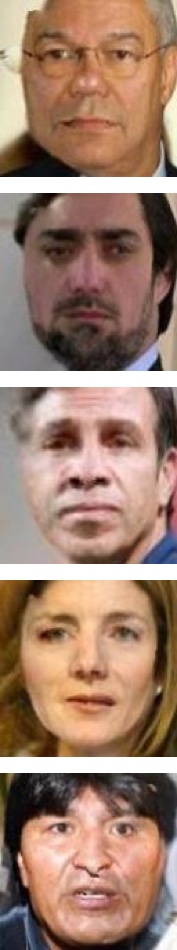}\vspace{0.03\linewidth}
	\end{minipage}}
	\subfigure[~\cite{hassner2015effective}]{
		\begin{minipage}[b]{0.13\linewidth}
			\centering
			\includegraphics[width=1\linewidth]{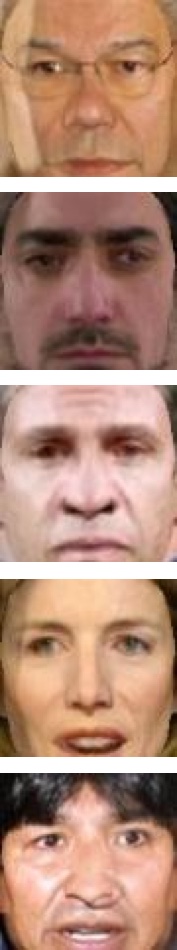}\vspace{0.03\linewidth}
	\end{minipage}}
	\caption{Face frontalization results on LFW datasets. Results of other methods are from~\cite{huang2017beyond}.}
	\label{fig:face_frontalization} 
	\vspace{-0.5cm}
\end{figure}

Figure~\ref{fig:seen_face_attributes_transfromation} presents the face synthesis results of the identities that appeare in the training dataset. Our method performs well in face synthesis, preserving both identity and attributes. 

Another important feature of our method is that it can synthesize unseen faces from the training set. Figure ~\ref{fig:unseen_face_attributes_transfromation} shows the zero-shot identity face synthesis results. Although our model never see these identities, but we can also generate high quality face images which keep the identity and attributes of the given faces.

In Figure~\ref{fig:face_frontalization}, we show that our framework can also be used for face frontalization. The results of other methods are from paper~\cite{huang2017beyond}. Unlike other methods our framework is not trained using pose annotations. With a frontal face as the input image to extract attributes, our framework can generate identity preserving frontal faces. Compared with~\cite{huang2017beyond}, the advantage of our method is that we are able to keep the lighting and skin color.

\begin{figure*}[t]
	\centering
	\includegraphics[width=1.8\columnwidth]{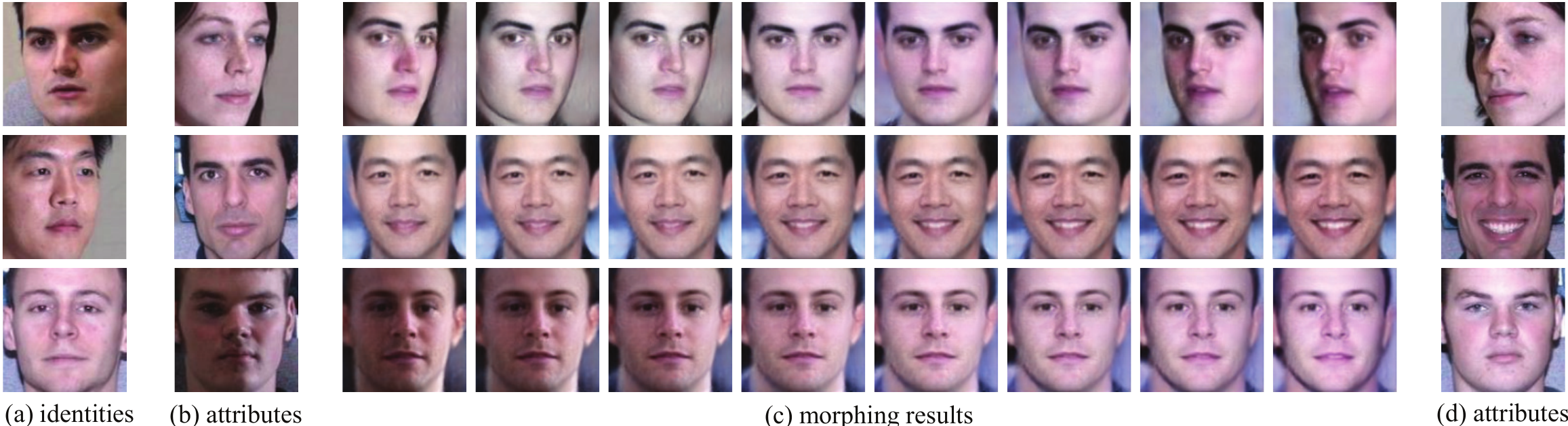}\\
	\caption{Face morphing results using unseen identities between two attributes. Our framework can gradually change pose, emotion, and lighting.}
	\label{fig:attributes_morphing}
	\vspace{-0.5cm}
\end{figure*}

\subsection{Face Attributes Morphing}
\label{sec:the_same_latent_vector_represent_same_attribute}

In this part, we validate that the attribute in the generated images will continuously change with the latent vector. We call this phenomenon attribute morphing. We test our model on the Multi-PIE~\cite{gross2010multi} dataset. We first select a pair of images $\bm{x}_1$ and $\bm{x}_2$, and then extract the attribute vector $\bm{z}_1$ and $\bm{z}_2$ using the attribute network $A$. Then, we obtain a series of attribute vectors $\bm{z}$ by linear interpolation, {\em i.e.}, $\bm{z} = \alpha \bm{z}_1 + (1 - \alpha)\bm{z}_2, \alpha \in [0, 1]$.
Figure~\ref{fig:attributes_morphing} presents the results of face attribute morphing. We can gradually change the pose, emotion, or lighting. For more results, please refer to the supplementary material.

\subsection{Face Adversarial Example Detection}

Deep network based face verification systems have been widely used in surveillance and access control. However, the existence of adversarial examples puts the security or safety of these systems at risk. In this experiment, we show that our framework can be used for face adversarial example detection without any modification.

For face verification, given two faces, we first extract the features for the two faces using a pretrained face classification DNN model. Then, we calculate the distance of the two features, and compare it to a threshold. If the feature distance is smaller than the threshold, they are predicted to have the same identity, and vice versa. 

Supposing two faces $\bm{x}_1$ and $\bm{x}_2$ have different identities, we can find imperceptible perturbations $\bm{r}$, such that $\bm{x}_1 + \bm{r}$ will be regarded as the same person as $\bm{x}_2$ using the above face verification system. Here, $\bm{x}_1 + \bm{r}$ is the adversarial sample. To find an adversarial sample, we optimize the problem.
\begin{equation}
\label{eqn:attack_VS}
\vspace{-0.3cm}
\begin{aligned}
\min& \|\bm{r}\|_2^2\\
s.t.& \|\bm{f}_C(\bm{x}_1 + \bm{r}) - \bm{f}_C(\bm{x}_2) \|_2^2 < \theta,
\end{aligned}
\end{equation}
Where $\bm{f}_C$ is the extracted feature from the pretrained network, and $\theta$ is the predefined threshold.

% To attack the model becomes clear. Let $\bm{x_s}$ is the source image of an adversarial example, $\bm{r}$ is the perturbations which will be added to the source image, $\bm{x_t}$ is the target image which the adversarial example wish to impersonate. 

% The extracted feature from the classification network $C$ for $\bm{x_s + r}$ and $\bm{x_t}$ are $f_C(x_s + r)$ and $f_C(x_t)$. So we define the optimization problem for attacking face verification system as:

% where
% \begin{displaymath}
% \label{eqn:cosine_distance}
% cos(\bm x, \bm y) = \frac{\bm x \cdot \bm y}{||\bm x|| \cdot ||\bm y||}
% \end{displaymath}

\begin{figure}[t]
	\centering
	\includegraphics[width=0.7\columnwidth]{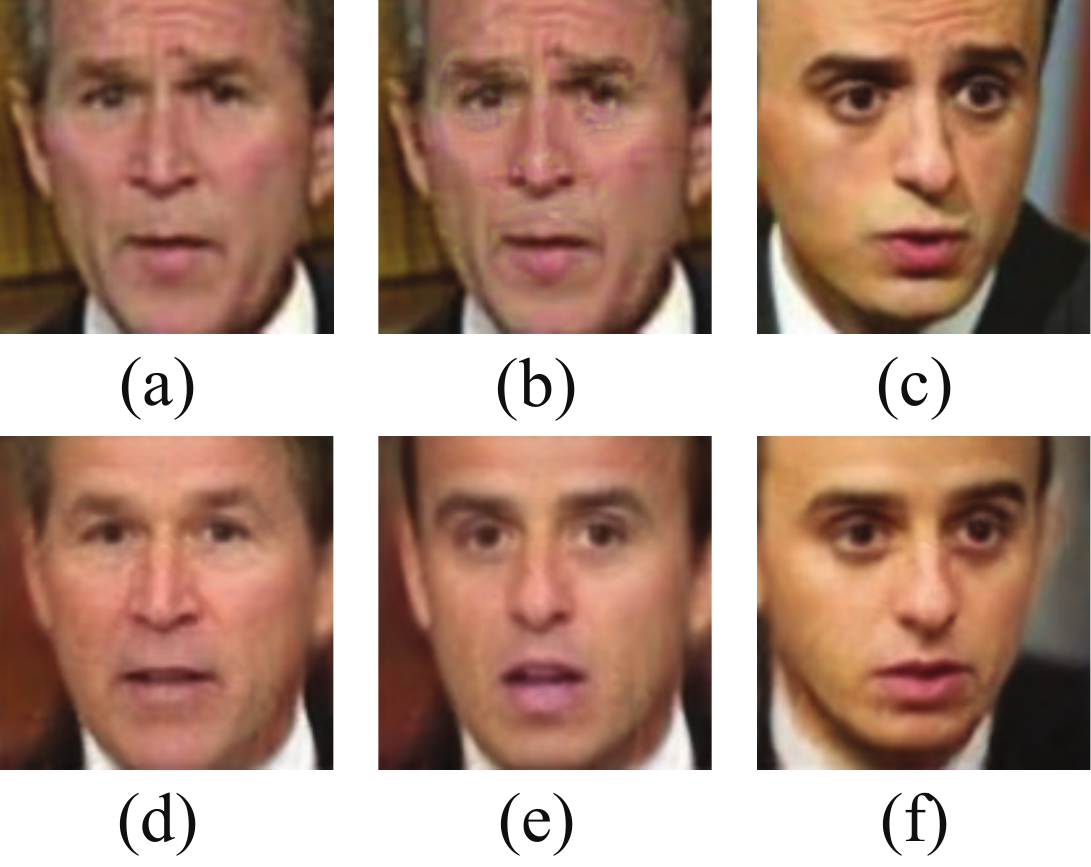}\\
	\caption{Adversarial example detection in face verification systems. (a) is the source image, (b) is the adversarial example which aims to attack face image (c).  (d), (e) and (f) are the reconstruction results from our framework. We can clearly observe that although the adversarial example shares a similar appearance with the source image, their reconstruction results have different appearances.}\label{fig:adversarial_attack}
	\vspace{-2mm}
\end{figure}

As shown in Figure~\ref{fig:adversarial_attack}, (a) and (c) are the two inputs $\bm{x}_1$ and $\bm{x}_2$, and (b) is the adversarial example $\bm{x}_1 + \bm{r}$. Since the adversarial examples have similar identity features with others faces, if we reconstruct the image from the feature using the proposed framework, it will generate an image of the other person. (e). The adversarial example and its reconstruction clearly have different identities. Based on this observation, we can use our generative model to reconstruct the faces, and compare the identity of the original faces and the reconstruction results to identify adversarial examples. 

We use the LFW~\cite{learned2016labeled} to conduct the experiments. For each of the $3000$ pairs of different identities, we generate two adversarial examples by performing adversarial attacks with each other. In total we obtain $6000$ adversarial examples for each predefined threshold. Here, we choose four different thresholds to conduct the experiments: $[0.4, 0.6, 0.8, 1]$. At the same time, we have $6000$ source images and their reconstruction. Although we can train another neural network to distinguish adversarial and source examples~\cite{metzen2017detecting}, the problem is that this neural network can be attacked again. Instead, we use the LBP~\cite{rahim2013face} feature. We extract LBP features from the input and its reconstruction image and then concatenate them. Finally, we train a linear SVM to conduct binary classification. The results are shown in Table~\ref{tab:adversarial_examples_detection_accu}, we can achieve $92.41\%$  accuracy if we require the feature distance to be less than $0.4$.

\begin{table}[t]
	\centering
	\begin{tabular}{p{2.0cm}|p{1.0cm}|p{1.0cm}|p{1.0cm}|p{1.0cm}}
		\hline
		threshold & 1.0 & 0.8 & 0.6 & 0.4  \\
		\hline
		acc     & 76.73\% &  82.58\%   &  87.18\%     & 92.41\%   \\
		\hline
	\end{tabular}
    \caption{Results of adversarial examples detection at different thresholds.}
	\label{tab:adversarial_examples_detection_accu}  
	\vspace{-2mm}
\end{table}

\section{Conclusion}
In this paper, we propose an Open-Set Identity Preserving Generative Adversarial Network framework for disentangling the identity and attributes of faces, synthesizing faces from the recombined identity and attributes. The framework shows superior performance in generating realistic and identity preserving face images, even for identities outside the training dataset. Our experiments demonstrate that our framework can also be applied to other tasks, such as face image frontalization, face attribute morphing, and adversarial example detection in face verification systems. In future work, we hope to explore whether our framework can be applied to other datasets, like birds, flowers, and rendered chairs. 

{\small
\bibliographystyle{ieee}
\bibliography{egbib}
}

\end{document}